\documentclass[10pt,twocolumn,letterpaper]{article}

\usepackage[pagenumbers]{wacv}

\usepackage[most]{tcolorbox}
\makeatletter
\newenvironment{figurehere}
  {\par\medskip\noindent\begin{minipage}{\linewidth}\def\@captype{figure}}
  {\end{minipage}\par\medskip}
\newenvironment{tablehere}
  {\par\medskip\noindent\begin{minipage}{\linewidth}\def\@captype{table}}
  {\end{minipage}\par\medskip}
\makeatother
\definecolor{rqblue}{HTML}{0F4D92}
\definecolor{rqteal}{HTML}{42949E}
\definecolor{rqviolet}{HTML}{9A4D8E}
\tcbset{
  rqbox/.style={
    enhanced,
    breakable,
    colback=#1!7,
    colframe=#1!75!black,
    boxrule=0.5pt,
    arc=1.5pt,
    left=5pt,
    right=5pt,
    top=3pt,
    bottom=3pt,
    before skip=6pt,
    after skip=6pt,
    fonttitle=\bfseries\sffamily,
    coltitle=black,
    attach title to upper={\par\medskip},
  },
}

\definecolor{wacvblue}{rgb}{0.21,0.49,0.74}
\usepackage[pagebackref,breaklinks,colorlinks,allcolors=wacvblue]{hyperref}

\def\wacvPaperID{22} 
\def\confName{WACV}
\def\confYear{2027}

\title{Beyond the Sink Interpretation: Radial Attention in Flow-Based VLA Policies}

\author{Yuxi Zheng \quad Yueqian Lin \quad Yudong Liu \quad Zijia Tang \\
Junyu Pan \quad Yiheng Wang \quad Hai Helen Li \quad Yiran Chen\thanks{Corresponding author.}\\
Duke University}

\begin{document}
\maketitle
\begin{abstract}
Prior work calls the high attention on the first and last action tokens of a flow-based VLA policy a radial action sink, and uses these tokens to set the execution horizon.
In language models, a sink is a token with high incoming attention, a small value, and little effect on the output.
We test this profile in the action expert of \(\pi_{0.5}\) on RoboCasa, with paired interventions that fix the observation and the initial noise.
The two endpoints receive 5.4\% of action-key mass, 1.4 percentage points above their uniform share, and rotary positional encoding produces this excess: directly in shallow layers and through upstream action-token RoPE deeper in the network.
The endpoints do not fit the sink profile.
Their value norms stay near the action-token mean in all 18 layers.
On 250 paired episodes, masking them or zeroing their values changes success by less than 4 points, and replacing their value directions with those of another observation lowers it by 3.2 to 7.2 points across three donors, whereas the same replacement on two non-endpoint tokens in the executed prefix lowers it by 28.4 points.
The token that does fit the sink profile on mass and value norm is the prefix BOS; zeroing its value raises success in two paired runs, by \(+13.2\) points (95\% CI \([+7.0,+19.4]\)) and \(+6.0\) points (\([+0.2,+11.8]\)).
Inside the action chunk, a relative-offset routing profile estimated on other tasks keeps 75\% of action-token spread and keeps closed-loop success on three held-out tasks (42.7\% against a 38.7\% baseline), while uniform or shuffled routing drops success to 3--4\%.
Attention mass reflects position; a token's function has to be tested by intervention.
\end{abstract}

\section{Introduction}
\label{sec:intro}

Flow-based vision-language-action policies denoise a chunk of continuous actions~\cite{black2025pi0,black2025pi05}.
In the action expert of \(\pi_{0.5}\), each action query attends, at each denoising step, to a multimodal prefix and to all action tokens.
Wang et al.~\cite{wang2026vla} observe that the first and last action tokens receive high incoming attention, call them radial action sinks, propose that they anchor temporal consistency across chunks, and use them to set an adaptive execution horizon.

The word \emph{sink} carries a specific profile from language models: one token receives much of the attention, its value is small, and reading it changes the output little~\cite{gu2025attention,barbero2025overmixing}.
If the endpoints had that profile, attention on them would park probability mass rather than carry action information.
We therefore ask three questions about the endpoints and the tokens around them.
Where does the endpoint mass come from?
Do the endpoint values behave like a sink's?
If not, which token does, and what keeps the action tokens distinct from each other?
The answers tell any method that reads these endpoints, including the horizon rule of~\cite{wang2026vla}, what kind of signal it is reading.

We answer these questions with paired interventions in the action expert of \(\pi_{0.5}\) on RoboCasa~\cite{robocasa2024}.
Each intervention and its baseline share the observation and the initial noise, and every closed-loop comparison is paired with the baseline of the same execution.

\begin{figurehere}
  \centering
  \includegraphics[width=\linewidth]{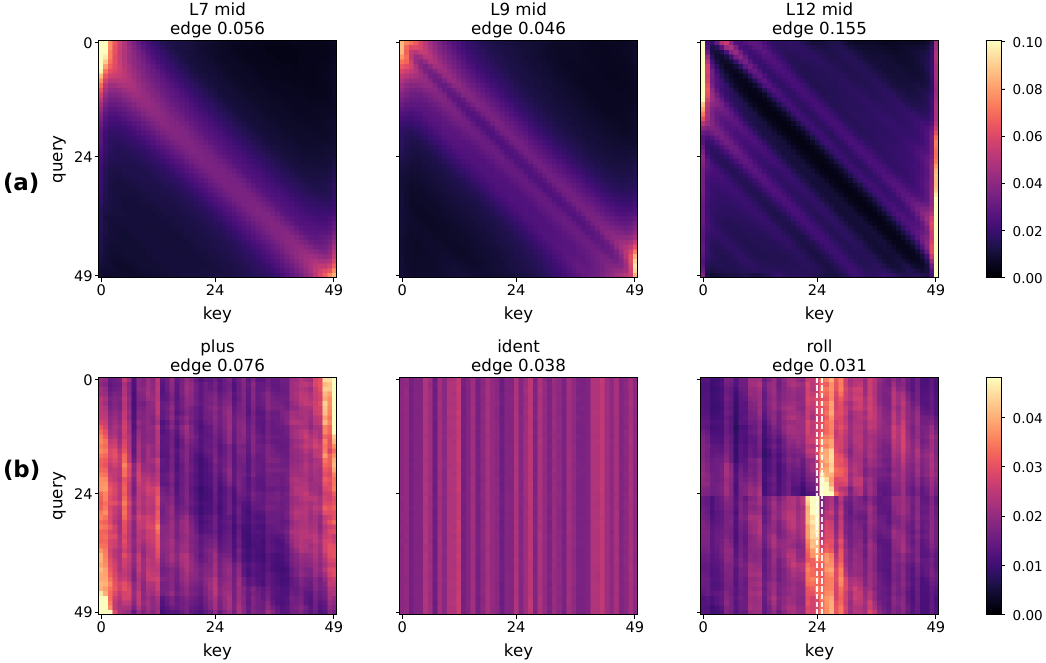}
  \caption{Endpoint attention is positional, and endpoint values are not inert.
(a) Mean action-key attention over \(1969\) chunks at layers 7, 9, and 12, mid noise.
(b) One chunk at layer 0, using the same pre-RoPE queries and keys: with RoPE (\texttt{plus}), without rotation (\texttt{ident}), and with shifted positions (\texttt{roll}).
The numbers printed in (b) are that chunk.}
  \label{fig:teaser}
\end{figurehere}

We find:
\begin{enumerate}
    \item \textbf{Endpoint mass is positional.} Current-layer RoPE produces it in shallow layers and upstream action-token RoPE in deeper layers; removing the relevant rotations returns it to the uniform reference at high and mid noise (Section~\ref{sec:radial}).
    \item \textbf{Endpoint values are not a null sink, and success is far less sensitive to them than to interior values.} Their norms and gated outputs are typical or larger. Masking or zeroing them changes closed-loop success by less than 4 points and replacing them by 3 to 7 points, while replacing two executed non-endpoint tokens lowers it by 25 to 28 points (Section~\ref{sec:sink}).
    \item \textbf{The sink-like token is the prefix BOS.} It has high mass and a small, shared value; zeroing that value raises success by \(+13.2\) and \(+6.0\) points in two runs (Section~\ref{sec:sink}).
    \item \textbf{Action routing follows a relative-offset profile.} A shared profile estimated on other tasks keeps 75\% of action-token spread and keeps success on three held-out tasks, while routings that break the offset structure drop success to 3--4\% (Section~\ref{sec:mixing}).
\end{enumerate}

\section{Background and Related Work}
\label{sec:background}

\subsection{Action Expert}
\label{sec:action-expert}

The action expert of $\pi_{0.5}$~\cite{black2025pi05} has 18 layers in the network we measure.
Each input is a multimodal prefix followed by a chunk of 50 action tokens, denoised jointly by flow matching.
The prefix begins with a \texttt{<bos>} token and then contains image tokens and language tokens.
Prefix queries read only the prefix, and do not read the action tokens.
Action queries read the prefix and every action key, so attention inside the action chunk is bidirectional.
Figure~\ref{fig:architecture} shows that head-averaged map at layer 2, with columns running from the prefix BOS through image and language into the action keys.
Queries and keys use rotary position embeddings~\cite{su2021roformer}, so changing a token's position changes its attention logits even when its content is held fixed.
Each layer scales the attention residual by an adaptive RMSNorm gate \(g\), conditioned on the flow-matching timestep, written \(g \odot W_O(AV)\).

\begin{figurehere}
  \centering
  \includegraphics[width=\linewidth]{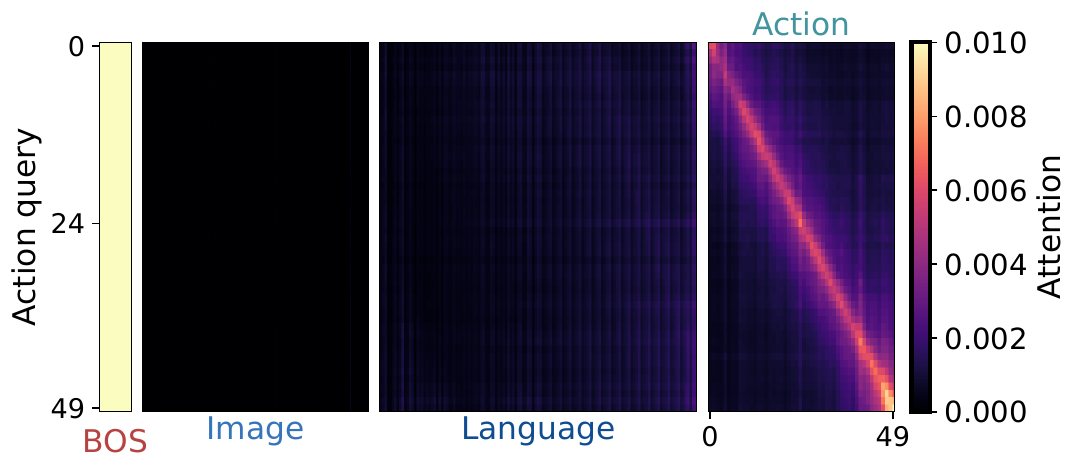}
  \caption{Head-averaged attention of action queries at layer 2.
  Columns are grouped into BOS, image, language, and action keys.
  The color scale stops at the brightest action-key entry.}
  \label{fig:architecture}
\end{figurehere}

\subsection{Related Work}
\label{sec:related}

Gu et al.~\cite{gu2025attention} treat an attention sink as a key-side bias on one early token: the sink key stores extra attention score, its value has a small \(\ell_2\) norm, and the attended token contributes little to the attention output.
Softmax ties the pattern to a competition among key scores that must sum to one~\cite{gu2025attention}.
A different internal token role appears in vision transformers.
Darcet et al.~\cite{darcet2024registers} find that low-information image tokens can be repurposed for internal computation, producing high-norm artifacts in low-information regions that are removed by adding register tokens.
This provides a distinct example of a token function beyond the low-contribution sink account.

Barbero et al.~\cite{barbero2025overmixing} argue that dumping attention onto the first token makes some heads nearly inactive, which slows over-mixing in deep, long-context language-model stacks.
They call the resulting update an approximate no-op: attention on the low-norm value keeps the attention output small, so the residual stream changes less.
These results describe one fixed early token in a causal language model.

RoPE encodes relative position through position-dependent rotations of the queries and the keys~\cite{su2021roformer}.
Chen et al.~\cite{chen2025hope} show that learned components inside RoPE can produce a global U-shaped pattern of attention.
These findings motivate separating positional concentration from the functional role of the attended token.

Diffusion transformers attend bidirectionally over the whole sequence, so high incoming mass is no longer tied to one early causal position.
Wu and Summa~\cite{wu2026attention} show that editing sink scores or sink values in text-to-image diffusion transformers shifts the generated image while text--image alignment holds.
Wang et al.~\cite{wang2026vla} observe the same kind of endpoint concentration on the first and last action tokens of a flow-based VLA, and use those tokens as anchors for the execution horizon.
We test the language-model sink profile on those endpoints: whether they have the small value norm and near-zero output contribution of a sink, and whether closed-loop control depends on reading them.
Action chunking and diffusion policies generate action sequences~\cite{zhao2023act,chi2023diffusionpolicy}, and VLA models such as RT-2 and OpenVLA map observations and instructions to actions~\cite{brohan2023rt2,kim2025openvla}. H{\"a}on et al.~\cite{haon2025steering} steer VLAs through directions in feedforward activations.
Our measurements stay on attention inside the action expert: where the endpoint mass comes from, whether the attended values are null, and which routing keeps the action tokens distinct.
\section{Paired Intervention Protocol}
\label{sec:framework}

\subsection{Experimental Setup}
\label{sec:setup}

We analyze the action expert of
$\pi_{0.5}$~\cite{black2025pi05} on
RoboCasa~\cite{robocasa2024}, using
the released checkpoint
\texttt{pi05\_pretrain\_human300} at step 75,000~\cite{robocasa365ckpt}.
Unless stated otherwise, we use the same five tasks
(CloseFridge, OpenCabinet, OpenDrawer,
PickPlaceCounterToCabinet, and TurnOnMicrowave),
environment seeds 0--2, and 10 denoising steps,
recording steps 0, 5, and 9, which we call high, mid, and last noise.

Interventions and baselines share the same observations
and the same initial noise.
RoPE experiments use separately collected rollout
chunks; the sample sizes differ with the episode lengths
and are reported in the corresponding table captions.
Other sampling and readout settings are specified
where applicable.

The incoming mass of key \(j\) averages attention onto that key,
\begin{equation}
  M_j=\frac{1}{N_qH}\sum_{i,h}A^{(h)}_{ij}.
  \label{eq:incoming-mass}
\end{equation}
\(A^{(h)}_{ij}\) is the attention from query \(i\) to key \(j\) in head \(h\), \(N_q\) is the number of queries, and \(H\) is the number of heads.
Incoming mass on a prefix key, including the \texttt{<bos>} key, uses this full softmax, so the prefix keys remain in the denominator.

Endpoint mass \(m_{\mathrm{edge}}\) is a different quantity.
Each head is first renormalized over the action keys only, and the heads are then averaged, so the prefix keys are not in the denominator.
For column \(c\), \(\bar A_c\) is the mean over queries more than one position away from \(c\), and
\begin{equation}
  m_{\mathrm{edge}}=\bar A_{0}+\bar A_{49}.
  \label{eq:m-edge}
\end{equation}
A uniform row gives \(m_{\mathrm{edge}}=2/50=0.040\).

Paired final-action deviation is the RMSE of an intervention's final chunk against the paired baseline chunk,
\begin{equation}
  D_{\mathrm{act}}=\mathrm{RMSE}(a_{\mathrm{intervened}},a_{\mathrm{baseline}}).
  \label{eq:d-act}
\end{equation}
\subsection{Positional Edits}
\label{sec:rope-intervention}

We first hold the pre-RoPE queries and keys fixed, so token content does not change,
and compare three positional transformations:
normal RoPE (\texttt{plus}), no rotation
(\texttt{ident}), and shifted positions (\texttt{roll}).
For action tokens at positions $P,\ldots,P+49$,
where $P$ is the prefix length, the normal and
no-rotation logits are
\begin{equation}
  L^+_{ij}
  = \frac{(R_{P+i}q_i)^\top(R_{P+j}k_j)}{\sqrt{d}},
  \qquad
  L^-_{ij}
  = \frac{q_i^\top k_j}{\sqrt{d}},
  \label{eq:rope-pair}
\end{equation}
where $R_p$ is the rotary transformation at position $p$.

The shifted condition assigns token $i$ the position
$P+((i+25)\bmod 50)$, placing the two extreme positions
on the middle tokens 24 and 25.
We compare mass on the original endpoint tokens, $m_{\mathrm{edge}}$,
with mass on the reassigned extreme positions,
$m_{\mathrm{geom}}$, to test whether the concentration
follows rotary position rather than token identity.

To test upstream dependence, we remove RoPE from
the earlier layers $0,\ldots,L-1$ and retain it at the measured layer $L$.
We then repeat the same three positional transformations
at layer $L$.
Unlike the current-layer edits, this upstream intervention
changes the representations that enter the measured layer.
Section~\ref{sec:deep-rope} further separates the
effects of upstream prefix- and action-token RoPE.

\subsection{Score and Value Edits}
\label{sec:score-value}

We separate changes to attention allocation from
changes to the attended values.
Score masking removes a selected key and renormalizes
the row over the remaining keys.
Value zeroing preserves those attention weights
but sets the selected value vector to zero.
Value replacement also preserves the weights.
At each edited position it copies the value of the same position from a reset observation of the same task, the donor, and rescales it to the recipient value's original norm, so only the direction changes.
The default donor is environment seed 99; seeds 100 and 101 serve as alternative donors.

These edits answer different questions.
Masking and zeroing test whether reading the value is required for the successes we measured.
Replacement tests whether a mismatched direction harms control.
To ask whether the endpoints are special, we apply the same replacement to two non-endpoint tokens, positions 2 and 3, and positions 2 and 4.
Each control cycle executes only the first five actions of the chunk (Appendix~\ref{app:protocol}).
Positions 2--4 are executed, token 0 is executed, and token 49 is not, so this control compares two executed interior tokens with one executed and one unexecuted endpoint.

For BOS replacement, we separately replace direction,
norm, or both: direction-only replacement preserves
the recipient norm, norm-only replacement preserves
its direction, and raw replacement changes both.
Section~\ref{sec:bos-information} evaluates these
interventions on 15 initial states.
\(D_{\mathrm{act}}\) and \(S(Z)\) use the 32 normalized
coordinates of the denoised chunk, before the output transform
(Appendix~\ref{app:protocol}).

Appendix~\ref{sec:bos-mask-gap} measures the local
effect of BOS masking after output projection and
timestep gating, with baseline queries, keys, and
values fixed.
Closed-loop endpoint interventions are detailed in
Appendix~\ref{app:endpoint-closed-loop}.
Closed-loop success is always compared, episode by episode, with the baseline of the same execution: GPU attention reductions are not deterministic, so two executions of the same policy disagree on 7 to 11\% of episode labels (Appendix~\ref{app:endpoint-closed-loop}).

\subsection{Action--Action Routing Edits}
\label{sec:mixing-intervention}

We use four interventions to test which properties of
action--action routing help preserve differences among
action tokens.
Each intervention changes only the weights on the
$N=50$ action keys.
For each layer and head, the prefix columns and each
query's total action-key mass
$m_i=\sum_{j=0}^{N-1}a_{ij}$ remain unchanged.
Section~\ref{sec:structured-mixing} reports the comparison;
per-permutation results are in Appendix~\ref{app:offset-profile}.

\paragraph{Uniform routing.}
The action-key mass is split equally, $a'_{ij}=m_i/N$.

\paragraph{Shared offset routing.}
Each action-key row is replaced by one offset profile,
estimated from unmodified forward passes and shared by
every query in that layer and head.
This profile is closer to the measured routing than uniform attention.
Normalize the row as $p_i(j)=a_{ij}/m_i$.
For query--key offset $\Delta=j-i$,
\begin{equation}
  T_{\ell,h}(\Delta)
  =
  \operatorname{mean}\{p_i(i+\Delta)\},
  \label{eq:offset-profile}
\end{equation}
where the mean runs over queries for which key $i+\Delta$ exists,
over the 10 denoising steps, and over the tasks used to estimate $T_{\ell,h}$.

The edited weights are
\begin{equation}
  a'_{ij}
  =
  m_i
  \frac{T_{\ell,h}(j-i)}
       {\sum_{k=0}^{N-1}T_{\ell,h}(k-i)},
  \label{eq:offset-routing}
\end{equation}
where the sum keeps only offsets that query $i$ can reach.
When $T_{\ell,h}$ is the mean of the same unmodified action-key rows, reconstructed
rows have mean cosine $0.831$ with those baseline rows
(Appendix~\ref{app:offset-profile}).

\paragraph{Offset shuffle.}
Replace $T_{\ell,h}(\Delta)$ in
Equation~\eqref{eq:offset-routing} by
$T_{\ell,h}(\pi(\Delta))$, where $\pi$ permutes the 99 offsets
and is shared across layers and heads.
The multiset of profile values is unchanged by that substitution.
Equation~\eqref{eq:offset-routing} then keeps only offsets the query can reach and renormalizes the row, so the edited row is not that same multiset, and the row entropy is not held fixed.

\paragraph{Key permutation.}
Form the aligned weights $a^T_{ij}$ from
Equation~\eqref{eq:offset-routing}, then set
$a^P_{i,\pi(j)}=a^T_{ij}$ with one permutation of the
50 action keys, shared across queries, layers, and heads.
Each row keeps its weights, and the matrix is no longer
a function of $j-i$.

\section{Endpoint Mass Comes from RoPE}
\label{sec:radial}

We first ask where the endpoint mass comes from, holding token content fixed and changing only rotary positions (Section~\ref{sec:rope-intervention}).

\subsection{Shallow Layers Follow the Current Rotation}
\label{sec:shallow-rope}

At layer 0, removing only the current rotation returns
endpoint mass to the uniform value of $0.040$ at all
three recorded noise levels. Reassigning the extreme
positions, with the same pre-RoPE content, moves the concentration to the tokens now at
those positions
(Table~\ref{tab:rope-pair},
Figure~\ref{fig:teaser}).
At layer 3, under high or mid noise, removing the current rotation
moves the ident mass toward that same uniform value, to \(0.046\) and \(0.047\)
(base rows of Supplementary Table~\ref{tab:rope-inherit-full}).
These layer-0 and layer-3 results show that RoPE at the current layer
can produce the endpoint concentration.

\subsection{Deep Layers Follow Upstream Action RoPE}
\label{sec:deep-rope}
Deep-layer endpoint concentration cannot be explained by the
rotation at the measured layer alone. At layer 12, the concentration
persists after the current rotation is removed
(Table~\ref{tab:rope-pair}), but falls to the
uniform reference of \(0.040\) at high and mid noise when
RoPE is also removed from the upstream layers \(0,\ldots,11\).
At the last step the same removal leaves the mass at
\(0.059\) to \(0.061\)
(Table~\ref{tab:rope-pos} and Supplementary Table~\ref{tab:rope-side}).
The same dependence on upstream RoPE appears where the unmodified ident mass is clearly above the uniform reference: layer 7 at mid noise, and layer 15 at high and mid noise.
After the upstream removal, ident mass is \(0.040\) in each of those three cases
(Supplementary Table~\ref{tab:rope-deep}).
Layer 16 shows the same high- and mid-noise drop
(Supplementary Table~\ref{tab:rope-inherit-full}).
Thus, at high and mid noise, the endpoint concentration reflects
positional processing accumulated in earlier layers, rather than only
the rotation applied at the layer being measured.

\begin{tablehere}
\caption{Endpoint mass at layer 0 and layer 12 under three rotations of the same pre-RoPE queries and keys.
The first three numeric columns are Equation~\eqref{eq:m-edge} after plus, ident, and roll, always on content tokens 0 and 49.
The last column is the same statistic on the two columns that roll assigns to the lowest and highest positions (24 and 25).
The uniform value is \(0.040\).
Means over 2,013 chunks from the five tasks in
Section~\ref{sec:setup}, environment seeds 0--2.
This is not the 2,134-chunk sample in
Supplementary Table~\ref{tab:rope-inherit-full}.}
\label{tab:rope-pair}
\centering
\footnotesize
\setlength{\tabcolsep}{3pt}
\begin{tabular}{@{}llcccc@{}}
\toprule
Noise & Layer & \(m^{+}\) & ident & roll & geom \\
\midrule
high & L0  & 0.072 & \textbf{0.040} & 0.032 & 0.070 \\
high & L12 & 0.133 & 0.069 & 0.048 & 0.074 \\
mid  & L0  & 0.076 & \textbf{0.040} & 0.033 & 0.073 \\
mid  & L12 & 0.157 & 0.154 & 0.094 & 0.061 \\
last & L0  & 0.060 & \textbf{0.040} & 0.036 & 0.056 \\
last & L12 & 0.101 & 0.107 & 0.080 & 0.049 \\
\bottomrule
\end{tabular}
\end{tablehere}

This upstream dependence is localized to the RoPE applied on the action tokens, not on the prefix.
Removing prefix RoPE alone does not remove the deep-layer concentration:
at mid noise, layer 12 stays near baseline (\(0.154\) against \(0.165\)) and layer 15 is only partially reduced (\(0.142\) against \(0.189\)), whereas removing upstream action-token RoPE returns both to \(0.040\) (Supplementary Table~\ref{tab:rope-side}).

Within the action chunk, RoPE on the two endpoint positions alone brings mid-noise mass from \(0.164\) to \(0.080\), and removing it from all action positions brings that mass to the uniform value
(Table~\ref{tab:rope-pos}).
Removing upstream RoPE only at positions 0 and 49 suppresses
concentration at high and mid noise, whereas the corresponding
edit at positions 24 and 25 leaves it near baseline.
Reassigning upstream action positions by a cyclic shift also
removes concentration from the original endpoints at high
and mid noise.
Together, these comparisons implicate the positional organization
of the 50-token action chunk, with a stronger contribution from
action-token RoPE than from prefix-token RoPE, and from the endpoint
positions than from the same number of central positions.

\begin{tablehere}
\caption{
Layer-12 endpoint mass after upstream action-token RoPE edits.
The forward pass keeps layer-12 RoPE.
The reported mass is recomputed at layer 12 without that rotation
(\texttt{ident}), so only the upstream edit varies.
All, Ends, and Center remove RoPE at all action positions,
positions 0 and 49, or positions 24 and 25.
Roll assigns token \(i\) the position
\(P + ((i+25)\bmod 50)\), where \(P\) is the prefix length.
Results use a separate sample of 2,046 chunks, not pooled
with Supplementary Tables~\ref{tab:rope-inherit-full}
and~\ref{tab:rope-side}.
The uniform reference is \(0.040\).
}
\label{tab:rope-pos}
\centering
\footnotesize
\setlength{\tabcolsep}{4pt}
\begin{tabular}{@{}lccccc@{}}
\toprule
Noise & Base & All & Ends & Center & Roll \\
\midrule
high & 0.065 & \textbf{0.040} & 0.033 & 0.065 & 0.035 \\
mid  & 0.164 & \textbf{0.040} & 0.080 & 0.154 & 0.036 \\
last & 0.112 & 0.061 & 0.146 & 0.107 & 0.080 \\
\bottomrule
\end{tabular}
\end{tablehere}
\section{Endpoints Do Not Match the Null-Sink Profile}
\label{sec:sink}

Section~\ref{sec:radial} traces the endpoint mass to rotary position.
That positional account does not settle whether the attended endpoint value is functionally null.
We therefore test the two action endpoints against the sink profile, first on their values and then in closed loop, and then test the shared prefix \texttt{<bos>} token.

\begin{tablehere}
\caption{Action-endpoint value norms stay near the action-token mean, and gated outputs stay above it.
$r_j$ divides the endpoint $\ell_2$ by the mean $\ell_2$ of the 50 action tokens at the same layer and denoising step, then averages over 18 layers ($n=1969$).
$c_j$ is Equation~\eqref{eq:gated-ratio}, averaged over 18 layers ($n=1955$).
High, mid, and last are denoising steps 0, 5, and 9.
The bottom row is the unweighted mean of those three steps.
The gated-output ratio is largest at high noise and closest to $1$ at the last step.}
\label{tab:sink-stats}
\centering
\footnotesize
\setlength{\tabcolsep}{4pt}
\begin{tabular}{@{}lcccc@{}}
\toprule
& \multicolumn{2}{c}{Value-norm ratio} & \multicolumn{2}{c}{Gated-output ratio} \\
\cmidrule(lr){2-3} \cmidrule(lr){4-5}
Noise & $r_0$ & $r_{49}$ & $c_0$ & $c_{49}$ \\
\midrule
high & 1.125 & 0.996 & 2.122 & 1.576 \\
mid  & 1.014 & 1.027 & 1.457 & 1.695 \\
last & 0.965 & 1.033 & 1.129 & 1.197 \\
3-step mean & 1.034 & 1.019 & 1.569 & 1.489 \\
\bottomrule
\end{tabular}
\end{tablehere}

\subsection{Endpoint Norms Stay Near the Action Mean}
\label{sec:endpoints}

The profile we use is incoming attention on one fixed token~\cite{xiao2024streaming}, a small value \(\ell_2\), and little contribution from that token to the attention output~\cite{gu2025attention}.
The action endpoints do not match that three-part profile.
We compare the two endpoints with this profile on two separate axes.

First, the action endpoints attract only a modest excess of
attention mass on average. Across all 18 layers and the three
recorded noise levels, the two endpoints together receive
5.4\% of the action-key attention mass, only 1.4 percentage points
above their uniform share of 4.0\%. 
The highest is \(11.7\) percentage points above that share, and the lowest is \(2.4\) below it (Figure~\ref{fig:medge-layers}).

\begin{figurehere}
  \centering
  \includegraphics[width=0.8\linewidth]{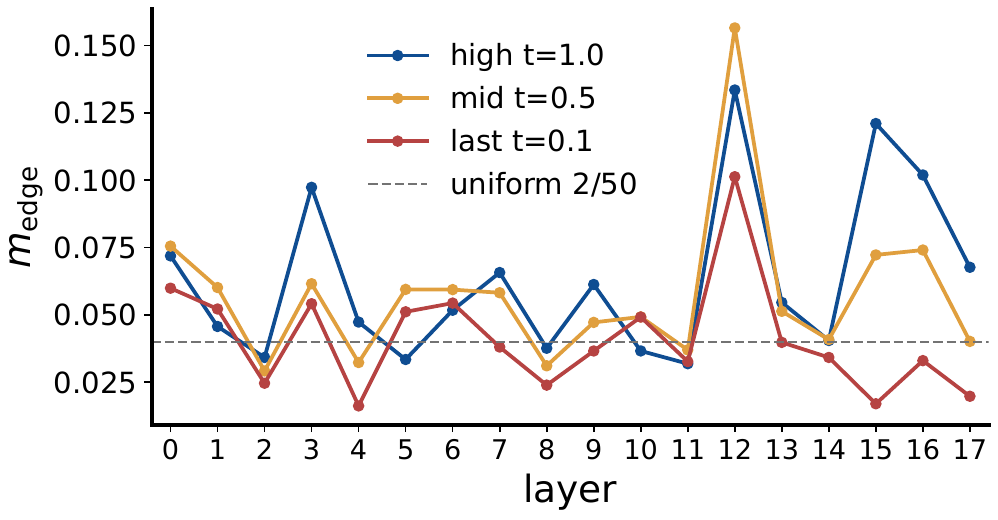}
  \caption{Unmodified endpoint mass across the 18 layers.
  \(m_{\mathrm{edge}}\) is the sum of the two endpoint columns after the query on that column and its two neighbors are dropped, as in Equation~\eqref{eq:m-edge}.
  The dashed line is the uniform value \(0.040\).
  Each curve is one recorded denoising step.}
  \label{fig:medge-layers}

\end{figurehere}
By comparison, Gu et al.~\cite{gu2025attention} report that the first token receives more than 30\% of attention on average in over 90\% of heads in several language models.

Second, endpoint values are neither unusually small nor
low-contributing.
Across 1,969 chunks, their norms relative to the action-token
mean average \(1.034\) and \(1.019\) at positions 0 and 49,
respectively, over all 18 layers and three recorded denoising steps
(Table~\ref{tab:sink-stats}).
To assess their contribution after output projection and
timestep gating, we measure the gated-output ratio
\begin{equation}
c_j
=
\frac{
  \operatorname{mean}_i\lVert g\odot W_O(A_{ij}V_j)\rVert
}{
  \operatorname{mean}_{k=0}^{49}
  \operatorname{mean}_i\lVert g\odot W_O(A_{ik}V_k)\rVert
}.
\label{eq:gated-ratio}
\end{equation}
Averaged over 18 layers and the three recorded denoising steps,
this ratio is \(1.569\) at position 0 and \(1.489\) at position 49
(Table~\ref{tab:sink-stats}).
\(c_j\) includes the attention weight, so it measures what each endpoint adds to the attention output, and both endpoints add more than the action-key mean.
Thus, unlike the small-value, low-contribution sink
profile reported for language models~\cite{gu2025attention}, the action endpoints have
typical value norms and above-average gated contributions.

\subsection{Success Is Insensitive to Endpoint Values}
\label{sec:endpoint-closed-loop}

Norms describe what the endpoints contribute, not whether the policy needs it, so we edit both endpoints, tokens 0 and 49, in every layer, and run 50 episodes per task on all five tasks.
All eight conditions run in one execution and are paired with its baseline, 113 of 250 (45.2\%).

With the default donor, none of the three endpoint edits changes success by more than 4 points (Table~\ref{tab:endpoint-success}).
Masking gives \(+3.2\) points (95\% CI \([-1.5,+7.9]\), \(p=0.24\)), zeroing \(+2.8\) (\([-2.7,+8.3]\), \(p=0.39\)), and replacement \(-3.6\) (\([-9.1,+1.9]\), \(p=0.25\)).
Aggregate success does not depend on reading the endpoint values; per-task results are in Appendix~\ref{app:endpoint-closed-loop}.

The same replacement on two executed non-endpoint tokens is far more harmful.
It lowers success by 28.4 points at positions 2 and 3 (\(p=2.3\times10^{-17}\)) and by 25.2 points at positions 2 and 4 (\(p=2.5\times10^{-13}\)), which is 24.8 and 21.6 points worse than endpoint replacement on the same episodes.

The gap does not follow edit size: with donors 100 and 101, endpoint replacement moves the executed actions more than interior replacement does (RMS 0.016 and 0.017, against 0.006), yet lowers success by only 7.2 and 3.2 points.
The three donors give \(-3.6\), \(-7.2\), and \(-3.2\) points; no donor lowers success by more than 7.2 points (donor-level tests and label flips are in Appendix~\ref{app:endpoint-closed-loop}).
An earlier execution of the same protocol gave \(-6.4\) points (\(p=0.036\)); its baseline and mask totals reproduce exactly, but 7--10\% of episode labels differ (Appendix~\ref{app:endpoint-closed-loop}).
All four replacement estimates are negative, so we read them as one small cost, 3 to 7 points, whose size varies with the donor and the execution.

Endpoint 0 is executed and endpoint 49 is not, while positions 2--4 are all executed.
Replacing endpoint directions costs a few points, and replacing executed interior directions costs a large share of all successes.

\begin{tablehere}
\caption{Closed-loop successes out of 50 when two action tokens are edited in every layer.
All columns come from one execution and share its baseline.
Replacement keeps the attention weights and the recipient norm and takes the direction from seed 99 of the same task.}
\label{tab:endpoint-success}
\centering
\footnotesize
\setlength{\tabcolsep}{2.5pt}
\begin{tabular}{@{}lcccccc@{}}
\toprule
& & \multicolumn{3}{c}{Endpoints (0, 49)} & \multicolumn{2}{c}{Interior} \\
\cmidrule(lr){3-5}\cmidrule(lr){6-7}
Task & Base & Mask & Value 0 & Repl. & (2, 3) & (2, 4) \\
\midrule
Fridge & 19 & 24 & 22 & 16 & 9 & 10 \\
Cabinet & 26 & 25 & 22 & 17 & 4 & 7 \\
Drawer & 30 & 32 & 33 & 37 & 28 & 32 \\
PickPlace & 34 & 33 & 35 & 26 & 1 & 1 \\
Microwave & 4 & 7 & 8 & 8 & 0 & 0 \\
\midrule
Total & 113 & 121 & 120 & 104 & \textbf{42} & \textbf{50} \\
\(\Delta\) (pp) & -- & \(+3.2\) & \(+2.8\) & \(-3.6\) & \(\mathbf{-28.4}\) & \(\mathbf{-25.2}\) \\
\bottomrule
\end{tabular}
\end{tablehere}

\subsection{The Prefix BOS Matches the Sink on Mass and Value Norm}
\label{sec:bos-mass}
The prefix \texttt{<bos>} is the token that matches the conventional sink on the two recorded axes: high incoming mass and a small value.
In 12 of 18 layers, \texttt{<bos>} receives more head-mean mass than either the image keys or the language keys (Figure~\ref{fig:bos-layers}, left).
Appendix~\ref{app:bos-closed-loop} shows that mass by head.
The largest gap is layer 1, where that mass is \(0.780\) against \(0.012\) on the image keys and \(0.112\) on the language keys.

The BOS value norm is below the image and language medians at every layer.
At layer 1 it is \(0.73\), against \(12.50\) and \(6.19\).
The nearest case is layer 0, \(10.64\) against an image median of \(20.43\).

The large \texttt{<bos>} entry in the residual-stream hidden state meets the massive-activation criterion of Sun et al.\ from layer 1 through layer 17~\cite{sun2024massive}, which has been associated with attention sink~\cite{queipo2026sinks}.
Its magnitude exceeds \(100\), and it is \(1{,}049\) to \(3{,}784\) times the median hidden-state entry.
This residual-stream magnitude is separate from the value-norm comparison above.

\begin{tablehere}
\caption{
BOS interventions on 15 initial states.
Value zero and cross-observation replacement preserve
attention weights; score mask removes the BOS key and
renormalizes attention.
\(D_{\mathrm{act}}\) measures final-action RMSE against
the paired baseline (Equation~\eqref{eq:d-act}).
}
\label{tab:bos-shared}
\centering
\footnotesize
\setlength{\tabcolsep}{4pt}
\begin{tabular}{@{}lcc@{}}
\toprule
Condition & Median \(D_{\mathrm{act}}\) & Mean \(D_{\mathrm{act}}\) \\
\midrule
value zero & \textbf{0.238} & \textbf{0.238} \\
score mask & 0.036 & 0.051 \\
replacement & 0.003 & 0.006 \\
\bottomrule
\end{tabular}
\end{tablehere}

\begin{figurehere}
  \centering
  \includegraphics[width=0.9\linewidth]{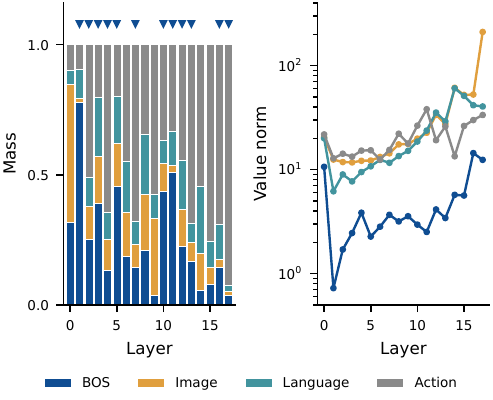}
  \caption{Left: head-mean mass on \texttt{<bos>}, on the image keys,
  on the language keys, and on the action keys.
  Triangles mark layers where the BOS mass exceeds both
  the image mass and the language mass.
  Right: value norms on a logarithmic axis.
  Layer indices are 0-based.}
  \label{fig:bos-layers}
\end{figurehere}

\subsection{The BOS Value Is Shared across Observations}
\label{sec:bos-information}

The \texttt{<bos>} value contributes to action prediction
but carries little observation-specific variation.
Across the 15 initial states, its direction and norm are
nearly invariant from layer 1 onward: median cross-observation
cosines range from \(0.955\) to \(0.999\), and the within-layer
maximum-to-minimum norm ratio has a median of \(1.024\).
We exclude layer 0, where this invariance follows trivially
from the fixed token embedding.

Cross-observation replacement tests whether these shared
values are functionally interchangeable.
Replacing the BOS value's direction, norm, or both yields
a median \(D_{\mathrm{act}}\leq 0.003\).
The row in Table~\ref{tab:bos-shared} is the raw replacement.
Zeroing the value with attention weights unchanged yields \(0.238\).
This contrast supports an approximately shared bias:
action prediction depends on the value's presence, but is
largely insensitive to its variation across observations.

Zeroing the BOS value raises closed-loop success from 107 to 140 of 250 episodes, \(+13.2\) points (95\% CI \([+7.0,+19.4]\), \(p<10^{-4}\); Table~\ref{tab:bos-closed-loop}).
A second paired run scales that value, \(v'=\lambda v\), with the attention weights and the key unchanged, and replicates the direction: zeroing (\(\lambda=0\)) gives \(+6.0\) points from a baseline of 114 (\([+0.2,+11.8]\), \(p=0.058\)), and halving (\(\lambda=0.5\)) gives \(+5.6\) points, so the gain does not require removing the value entirely.
On seeds where both conditions succeed, value zero also finishes sooner: 393 against 443 environment steps over 94 pairs.
Appendix~\ref{app:bos-closed-loop} gives the per-task counts, the paired tests, and the variation between executions.

\begin{tablehere}
\caption{Closed-loop success when the prefix \texttt{<bos>} token is edited.
Each cell is successes out of 50 episodes.
The total is out of 250.
Value zero has a paired difference of \(+13.2\) points (Wald 95\% CI \([+7.0,+19.4]\)).
Score masking is \(+3.6\) points (\([-1.5,+8.7]\)).
Replacement is \(0.0\) points (\([-4.6,+4.6]\)).}
\label{tab:bos-closed-loop}
\centering
\footnotesize
\setlength{\tabcolsep}{3pt}
\begin{tabular}{@{}lcccc@{}}
\toprule
Task & Base & Mask & Value 0 & Replacement \\
\midrule
OpenDrawer & 28 & 29 & 35 & 31 \\
OpenCabinet & 23 & 24 & 30 & 23 \\
CloseFridge & 19 & 22 & 33 & 17 \\
PickPlace & 32 & 35 & 32 & 31 \\
TurnOnMicrowave & 5 & 6 & 10 & 5 \\
\midrule
Total & 107 & 116 & \textbf{140} & 107 \\
\bottomrule
\end{tabular}
\end{tablehere}

\section{Action Spread Follows the Offset Profile}
\label{sec:mixing}

Attention sinks in causal language models have been proposed
to limit over-mixing among the content tokens
\cite{barbero2025overmixing}.
Section~\ref{sec:sink} shows that the prefix
\texttt{<bos>} value acts as an approximately shared bias across observations.
We now ask whether attention onto that BOS token also helps preserve
differences among the 50 action tokens.

We measure differences among tokens using their
root-mean-square distance from the chunk mean.
For a chunk of vectors $Z=(z_0,\ldots,z_{N-1})$,
the spread is
\begin{equation}
  S(Z)
  =
  \left(
    \frac{1}{N}
    \sum_{i=0}^{N-1}
    \|z_i-\bar z\|_2^2
  \right)^{1/2},
  \qquad
  \bar z=\frac{1}{N}\sum_{i=0}^{N-1}z_i,
  \label{eq:token-spread}
\end{equation}
where $N=50$.

For the final denoised action chunk $A$, we report
the spread ratio:
\begin{equation}
  r_{\mathrm{spread}}
  =
  \frac{S(A^{\mathrm{edit}})}
       {S(A^{\mathrm{base}})}.
  \label{eq:relative-spread}
\end{equation}
The paired baseline uses the same observation and the same initial noise
as the intervention.
A spread ratio of one indicates that the chunk spread is unchanged;
a ratio near zero indicates nearly identical action vectors
within the chunk.

On 24 observation--noise pairs from three tasks, masking the BOS score leaves the final-chunk spread ratio at \(0.987\), and zeroing the BOS value raises it to \(1.381\) (Supplementary Table~\ref{tab:bos-spread}).
Attention to the BOS key is therefore not what preserves action-token spread.

\subsection{A Shared Offset Routing Retains Spread}
\label{sec:structured-mixing}

We apply shared offset routing and uniform routing,
as defined in Section~\ref{sec:mixing-intervention}, to all 18 layers.
For each held-out task, \(T_{\ell,h}\) is estimated on the other two
tasks and then tested on that held-out task, using 4 observations and 2 noises.

The mean spread ratio of shared offset routing is \(0.752\),
against \(0.024\) for uniform routing on the same pairs
(Table~\ref{tab:offset-profile}).
Across the held-out tasks it ranges from \(0.645\) to \(0.845\).
TurnOnMicrowave is the lowest, at \(0.645\), and it remains
above that shared routing's uniform ratio of \(0.015\).
The mean \(D_{\mathrm{act}}\) is \(0.164\).

Shared offset routing removes all differences across observations and queries, keeping one relative-offset profile per layer and head.
Offset shuffle keeps the multiset of those profile values but assigns them to different offsets, and the spread ratio falls to \(0.194\).
Row entropy under the shuffle stays within \(0.12\) nats of shared offset routing (\(3.39\)--\(3.41\) against \(3.51\) nats, Appendix~\ref{app:offset-profile}), so the drop to \(0.194\) comes from the arrangement of the weights, not from a change in row sharpness.
Key permutation keeps the weights inside each row but assigns them to different action keys, and the row is no longer a function of \(j-i\).
The spread ratio falls further, to \(0.096\).
The arrangement of the weights over offsets, not the set of weight values, is what retains the spread.

\begin{tablehere}
\caption{
Leave-one-task-out offset routing on all 18 layers.
\(T_{\ell,h}\) for each held-out task is estimated on the other two.
Each test fold uses 4 observations and 2 noises.
Spread ratios use Equation~\eqref{eq:relative-spread}.
Permutation entries are means over five fixed permutations
(Appendix~\ref{app:offset-profile}).
}
\label{tab:offset-profile}
\centering
\small
\begin{tabular}{lcc}
\toprule
Condition & Spread ratio & \(D_{\mathrm{act}}\) \\
\midrule
baseline & 1.000 & 0.000 \\
shared offset routing & 0.752 & 0.164 \\
offset shuffle & 0.194 & 0.427 \\
key permutation & 0.096 & 0.430 \\
uniform & 0.024 & 0.438 \\
\bottomrule
\end{tabular}
\end{tablehere}

\subsection{The Offset Profile Keeps Closed-Loop Success}
\label{sec:route-closed-loop}

Spread describes the chunk, not the task.
We therefore run the four routing edits in closed loop on the three tasks above, each with the profile estimated on the other two, for 50 episodes per task, paired with the baseline of the same execution (Table~\ref{tab:route-closed-loop}).

Shared offset routing keeps success: 64 of 150 against 58, \(+4.0\) points (95\% CI \([-2.9,+10.9]\), \(p=0.35\)).
Offset shuffle, key permutation, and uniform routing each drop success to 5 or 6 of 150, a loss of 34.7 to 35.3 points (\(p<10^{-13}\)).
Paired against those three, shared offset routing is higher by 38.7 to 39.3 points.
Shared offset routing keeps both spread and success, and the three edits that cut spread below \(0.2\) of baseline all cut success to 3--4\%; among those three, success does not track the remaining differences in spread.
The two readouts use different samples, 24 observation--noise pairs and 150 episodes, so they co-vary; one is not a measurement of the other.

The profile is estimated per task family; Appendix~\ref{app:offset-profile} reports the two remaining tasks, OpenCabinet and PickPlace.

\begin{tablehere}
\caption{Closed-loop successes out of 50 under all-layer routing edits.
Each task uses the profile estimated on the other two.
All rows come from one execution and share its baseline.}
\label{tab:route-closed-loop}
\centering
\footnotesize
\setlength{\tabcolsep}{3pt}
\begin{tabular}{@{}lcccc@{}}
\toprule
Condition & Drawer & Microwave & Fridge & Total (\%) \\
\midrule
baseline & 33 & 4 & 21 & 58 (38.7) \\
shared offset & 37 & 5 & 22 & \textbf{64 (42.7)} \\
offset shuffle & 4 & 0 & 1 & 5 (3.3) \\
key permutation & 4 & 0 & 2 & 6 (4.0) \\
uniform & 4 & 0 & 2 & 6 (4.0) \\
\bottomrule
\end{tabular}
\end{tablehere}

\section{Discussion and Limitations}
\label{sec:discussion}
\label{sec:limitations}

Endpoint attention in this policy is a property of position, not of the tokens that occupy it.
The endpoint values make normal-sized, non-null contributions, yet closed-loop success depends on them far less than on executed interior tokens, and the token that matches the sink profile is the prefix BOS.
A horizon rule can still read the endpoint pattern as an anchor; our results identify that anchor as a RoPE-induced position signal rather than a low-contribution sink.
The same distinction applies to any method that selects tokens by attention mass, for anchoring, pruning, or cache compression: in this policy the chunk ends draw extra mass from position alone.
Moving positions with content fixed, then masking, zeroing, and replacing values against a non-endpoint control, separates position from function.

These measurements come from one \(\pi_{0.5}\) checkpoint with bidirectional action attention and a shared prefix token.
SmolVLA~\cite{shukor2025smolvla}, with causal action attention and no BOS token, shows no endpoint concentration (Appendix~\ref{app:smolvla}), so the endpoint signature belongs to this attention topology.
The endpoint--interior contrast pairs one executed and one unexecuted endpoint with two executed interior tokens, and the offset profile is estimated per task family on three held-out tasks.
On this checkpoint the BOS scaling factor tracks action magnitude (\(+13.2\) and \(+6.0\) points in two runs), which makes zeroing the BOS value a concrete first test on any other flow-based VLA.

\section{Conclusion}
\label{sec:conclusion}

In the action expert of \(\pi_{0.5}\)~\cite{black2025pi05}, RoPE produces the radial endpoint attention, but the endpoints are not null sinks: editing their values moves success by at most 7 points, against 25 to 28 for executed interior tokens.
The sink-like token is the prefix \texttt{<bos>}, and action routing works through a relative-offset profile.
Attention mass reflects position; a token's role has to be tested by intervention.

\section*{Acknowledgments}
This work was supported in part by NSF (2625580), ARO (W911NF-23-2-0224), and DOE (DE-SC0026696).

{
    \small
    \bibliographystyle{ieeenat_fullname}
    \bibliography{main}
}
\clearpage
\appendix

\section{Evaluation Protocol}
\label{app:protocol}

The sections below follow the main text.
Open-loop \(D_{\mathrm{act}}\) and \(S(Z)\) use the denoised model tensor: 50 steps by 32 normalized coordinates, before the output transform.
The environment receives the first 12 coordinates after that transform.
Closed-loop episodes use 50 resets per task, seeds 0--49, on the five tasks in Section~\ref{sec:setup}.
Seeds 0--2 are the first three of those resets.
The open-loop tables that cite seeds 0--2 use only those three.
Episode caps are 900, 1050, 750, 750, and 450 environment steps for CloseFridge, OpenCabinet, OpenDrawer, PickPlaceCounterToCabinet, and TurnOnMicrowave.
Each cycle denoises a chunk of 50 actions, executes the first 5, and then replans.
Success ends the episode.

\section{RoPE Instrument Validation}
\label{app:rope}

Table~\ref{tab:rope-inherit-full} is the per-layer skip used to test upstream RoPE.
Base keeps RoPE on the upstream layers.
Skip removes RoPE on layers \(0,\ldots,L-1\) and keeps RoPE at layer \(L\).
Layer 0 and layer 12 use the 2134-chunk sample that skips layers 0--11.
Layers 3 and 16 use the 2173-chunk sample that skips \(0,\ldots,L-1\) for each target.
The two samples are not pooled.
Ident mass for the clear high- and mid-noise drops, layer 7 at mid noise and layer 15 at high and mid noise, is reported in Table~\ref{tab:rope-deep}.
Layer 0 is a sanity check: it has no upstream layers, so base and skip coincide.
Layer 3 at high and mid noise keeps plus after the skip, which matches the current-layer RoPE contrast at layer 3 in the main paper.

\begin{tablehere}
\caption{Endpoint mass before and after removing upstream RoPE.
Columns match Table~\ref{tab:rope-pair}.
Layer 0 and layer 12: 2134 chunks, skip layers 0--11.
Layers 3 and 16: 2173 chunks, skip layers \(0,\ldots,L-1\).
The clear ident drops at layers 7 and 15 are in Table~\ref{tab:rope-deep}.}
\label{tab:rope-inherit-full}
\centering
\scriptsize
\setlength{\tabcolsep}{2pt}
\begin{tabular}{@{}llccccc@{}}
\toprule
Noise & Layer & Pass & \(m^{+}\) & ident & roll & geom \\
\midrule
high & L0  & base & 0.072 & 0.040 & 0.032 & 0.071 \\
high & L0  & skip & 0.072 & 0.040 & 0.032 & 0.071 \\
high & L3  & base & 0.095 & 0.046 & 0.037 & 0.083 \\
high & L3  & skip & 0.092 & 0.040 & 0.030 & 0.090 \\
high & L12 & base & 0.126 & 0.068 & 0.047 & 0.070 \\
high & L12 & skip & 0.086 & 0.040 & 0.031 & 0.086 \\
high & L16 & base & 0.114 & 0.190 & 0.113 & 0.050 \\
high & L16 & skip & 0.062 & 0.040 & 0.042 & 0.063 \\
\midrule
mid  & L0  & base & 0.075 & 0.040 & 0.033 & 0.072 \\
mid  & L0  & skip & 0.075 & 0.040 & 0.033 & 0.072 \\
mid  & L3  & base & 0.061 & 0.047 & 0.042 & 0.053 \\
mid  & L3  & skip & 0.061 & 0.042 & 0.036 & 0.056 \\
mid  & L12 & base & 0.160 & 0.162 & 0.102 & 0.057 \\
mid  & L12 & skip & 0.068 & 0.041 & 0.037 & 0.068 \\
mid  & L16 & base & 0.073 & 0.186 & 0.076 & 0.043 \\
mid  & L16 & skip & 0.043 & 0.040 & 0.044 & 0.044 \\
\midrule
last & L0  & base & 0.060 & 0.040 & 0.036 & 0.056 \\
last & L0  & skip & 0.060 & 0.040 & 0.036 & 0.056 \\
last & L3  & base & 0.053 & 0.050 & 0.047 & 0.039 \\
last & L3  & skip & 0.055 & 0.054 & 0.048 & 0.039 \\
last & L12 & base & 0.105 & 0.112 & 0.085 & 0.047 \\
last & L12 & skip & 0.045 & 0.058 & 0.049 & 0.035 \\
last & L16 & base & 0.033 & 0.060 & 0.039 & 0.031 \\
last & L16 & skip & 0.022 & 0.052 & 0.033 & 0.019 \\
\bottomrule
\end{tabular}
\end{tablehere}

\begin{tablehere}
\caption{Ident endpoint mass where removing upstream RoPE returns the mass to the uniform reference.
Skip removes RoPE on layers \(0,\ldots,L-1\) and keeps it at layer \(L\).
The uniform reference is \(0.040\).
Means over 2,173 chunks.}
\label{tab:rope-deep}
\centering
\footnotesize
\setlength{\tabcolsep}{4pt}
\begin{tabular}{@{}llcc@{}}
\toprule
Noise & Layer & Base & Skip \\
\midrule
high & L15 & 0.155 & \textbf{0.040} \\
mid  & L7  & 0.088 & \textbf{0.040} \\
mid  & L15 & 0.175 & \textbf{0.040} \\
\bottomrule
\end{tabular}
\end{tablehere}

Reversal assigns the positions \(P+49,\ldots,P\).
Every relative offset changes sign, and the lowest and highest positions remain on content tokens 0 and 49, with the two ends exchanged.
Table~\ref{tab:rope-rev} reports \(m_{\mathrm{edge}}\) on those tokens.
At layer 0, the change from plus is small.
At layer 12 and high noise, reversal falls to \(0.068\), beside ident at \(0.069\).

\begin{tablehere}
\caption{Endpoint mass under position reversal, on the sample of Table~\ref{tab:rope-pair}.
Rev is \(m_{\mathrm{edge}}\) on content tokens 0 and 49.}
\label{tab:rope-rev}
\centering
\footnotesize
\setlength{\tabcolsep}{4pt}
\begin{tabular}{@{}llcccc@{}}
\toprule
Noise & Layer & \(m^{+}\) & ident & rev & roll \\
\midrule
high & L0  & 0.072 & 0.040 & 0.072 & 0.032 \\
high & L3  & 0.097 & 0.046 & 0.083 & 0.037 \\
high & L12 & 0.133 & 0.069 & 0.068 & 0.048 \\
mid  & L0  & 0.076 & 0.040 & 0.076 & 0.033 \\
mid  & L3  & 0.062 & 0.047 & 0.058 & 0.042 \\
mid  & L12 & 0.157 & 0.154 & 0.118 & 0.094 \\
last & L0  & 0.060 & 0.040 & 0.060 & 0.036 \\
last & L3  & 0.054 & 0.051 & 0.055 & 0.049 \\
last & L12 & 0.101 & 0.107 & 0.092 & 0.080 \\
\bottomrule
\end{tabular}
\end{tablehere}

\begin{tablehere}
\caption{
Endpoint mass measured without the target layer's rotation
(\texttt{ident}) after upstream RoPE interventions.
Prefix, Action, and Both remove RoPE from earlier layers
on prefix tokens, action tokens, or both, respectively.
Base uses unmodified upstream layers.
Results use 2,062 chunks; the uniform reference is \(0.040\).
}

\label{tab:rope-side}
\centering
\footnotesize
\setlength{\tabcolsep}{4pt}
\begin{tabular}{@{}llcccc@{}}
\toprule
Noise & Layer & Base & Prefix & Action & Both \\
\midrule
high & L12 & 0.066 & 0.060 & \textbf{0.040} & \textbf{0.040} \\
mid  & L12 & 0.165 & 0.154 & \textbf{0.040} & \textbf{0.040} \\
last & L12 & 0.115 & 0.118 & 0.060 & 0.059 \\
high & L15 & 0.162 & 0.092 & \textbf{0.040} & \textbf{0.040} \\
mid  & L15 & 0.189 & 0.142 & \textbf{0.040} & \textbf{0.040} \\
\bottomrule
\end{tabular}
\end{tablehere}

\subsection{Side and Position Edit Checks}
\label{app:rope-side}

Table~\ref{tab:rope-side} accompanies Section~\ref{sec:deep-rope}, and Table~\ref{tab:rope-pos} is reported there.
On the side sample, recomputed plus attention at the scored layer matches the captured attention, with Pearson 1.000 at the reported precision.
Layer 0 endpoint mass matches the unedited pass under every side edit; the absolute difference is below \(10^{-6}\).
On the position sample, layer 0 plus is unchanged under every edit.
The full action-token removal also changes the computation that reaches layer 12.
At mid noise, its pre-RoPE queries and keys differ from the paired baseline by RMSE \(0.550\), against \(0.170\) for the endpoint edit and \(0.129\) for the center edit.
\(D_{\mathrm{act}}\) is \(0.298\), \(0.098\), and \(0.057\) for those three edits.
Roll returns ident to \(0.035\) at high noise and \(0.036\) at mid noise, but the RMSE of its pre-RoPE queries and keys is \(0.414\) at mid noise and its \(D_{\mathrm{act}}\) is \(0.314\).

\section{Closed-Loop Endpoint Interventions}
\label{app:endpoint-closed-loop}

Two tokens are edited together at every layer.
All eight conditions run in one execution: five tasks, seeds 0--49, a replan of 5 steps, 10 denoising steps, and the policy-noise key \(\mathrm{seed}\cdot10007+\mathrm{chunk}+17\).
Score masking drops the edited keys and renormalizes the remaining keys.
Value zero keeps the weights and sets the edited values to zero.
Replacement keeps the weights and inserts, at each edited position, the value of the same position from the donor, a reset observation of the same task at seed 99, 100, or 101, rescaled to the recipient norm.
Table~\ref{tab:endpoint-paired} gives the paired difference, intervention minus baseline, with a Wald matched-pairs 95\% interval and the exact two-sided McNemar \(p\).
The \(p\)-values are not corrected for the seven comparisons.

\begin{tablehere}
\caption{Paired success changes against the baseline of the same execution, 113 of 250.
\(s\!\to\!f\) is success to failure, and \(f\!\to\!s\) the reverse.
Differences are in percentage points.}
\label{tab:endpoint-paired}
\centering
\scriptsize
\setlength{\tabcolsep}{2pt}
\begin{tabular}{@{}lrrrrl@{}}
\toprule
Edit & Succ. & \(s\!\to\!f\) & \(f\!\to\!s\) & Diff. [95\% CI] & \(p\) \\
\midrule
Mask (0, 49) & 121 & 14 & 22 & \(+3.2\) \([-1.5,+7.9]\) & 0.24 \\
Value 0 (0, 49) & 120 & 21 & 28 & \(+2.8\) \([-2.7,+8.3]\) & 0.39 \\
Repl.\ (0, 49), d99 & 104 & 29 & 20 & \(-3.6\) \([-9.1,+1.9]\) & 0.25 \\
Repl.\ (0, 49), d100 & 95 & 36 & 18 & \(-7.2\) \([-12.9,-1.5]\) & 0.020 \\
Repl.\ (0, 49), d101 & 105 & 30 & 22 & \(-3.2\) \([-8.8,+2.4]\) & 0.33 \\
Repl.\ (2, 3), d99 & 42 & 76 & 5 & \(-28.4\) \([-34.5,-22.3]\) & \(2.3{\times}10^{-17}\) \\
Repl.\ (2, 4), d99 & 50 & 72 & 9 & \(-25.2\) \([-31.5,-18.9]\) & \(2.5{\times}10^{-13}\) \\
\bottomrule
\end{tabular}
\end{tablehere}

Paired against endpoint replacement with the same donor, replacement at (2, 3) is lower by 24.8 points (\([+19.1,+30.5]\), \(p=3.6\times10^{-16}\)) and at (2, 4) by 21.6 points (\([+16.0,+27.2]\), \(p=2.6\times10^{-13}\)).
Mask and value zero differ by 0.4 points (\(p=1.0\)).

\paragraph{Per task.}
Endpoint replacement with donor 99 changes success by \(-6\) (Fridge), \(-18\) (Cabinet, \(p=0.035\)), \(+14\) (Drawer, \(p=0.016\)), \(-16\) (PickPlace, \(p=0.022\)), and \(+8\) (Microwave) points.
The effect is heterogeneous in sign, so it cannot be read as a uniform loss of information.
Mask and value zero move the five tasks in mixed directions, between \(-8\) and \(+10\) points, with no task at \(p<0.15\).
Interior replacement at (2, 3) changes them by \(-20\), \(-44\), \(-4\), \(-66\), and \(-8\) points; Microwave starts at 4 of 50 and cannot fall further than 8 points.

\paragraph{Donor and execution variation.}
On the same 250 episodes, donors 99 and 100 give different success labels on 9.2\% of episodes, 99 and 101 on 10.8\%, and 100 and 101 on 8.0\%.
The first four conditions repeat an earlier execution of the same protocol: totals match for baseline (113) and mask (121) and differ by 2 and 7 for value zero and replacement, with 90.4--93.2\% of episode labels in agreement.
The seed fixes the environment and the diffusion noise but not the order of GPU attention reductions.
That earlier execution gave value zero \(+2.0\) points and replacement \(-6.4\) points (\(p=0.036\)) against the same baseline count.

\paragraph{Edit size.}
Table~\ref{tab:endpoint-strength} measures how far each edit moves the five executed actions from the paired baseline on fixed observations.
Endpoint replacement with donors 100 and 101 moves them more than interior replacement, yet harms success less.
The median change in executed-action smoothness is below \(2\times10^{-4}\) for every edit, so the interior losses do not come from jitter.
Every edit changes the action on a fixed observation by more than \(10^{-3}\), and the baseline reproduces with zero difference.

\begin{tablehere}
\caption{RMS difference of the five executed actions from the paired baseline, averaged over the five tasks, with the range across tasks.}
\label{tab:endpoint-strength}
\centering
\footnotesize
\setlength{\tabcolsep}{4pt}
\begin{tabular}{@{}lcc@{}}
\toprule
Edit & Mean RMS & Range \\
\midrule
Mask (0, 49) & 0.0107 & \([0.0031, 0.0192]\) \\
Value 0 (0, 49) & 0.0210 & \([0.0114, 0.0370]\) \\
Repl.\ (0, 49), d99 & 0.0027 & \([0.0006, 0.0059]\) \\
Repl.\ (0, 49), d100 & 0.0156 & \([0.0012, 0.0714]\) \\
Repl.\ (0, 49), d101 & 0.0171 & \([0.0007, 0.0784]\) \\
Repl.\ (2, 3), d99 & 0.0063 & \([0.0041, 0.0091]\) \\
Repl.\ (2, 4), d99 & 0.0062 & \([0.0029, 0.0101]\) \\
\bottomrule
\end{tabular}
\end{tablehere}

\section{Local Effect of Masking BOS}
\label{sec:bos-mask-gap}

Score masking produces a smaller denoised-action deviation than
value zeroing on all 15 initial states under paired noise
(Table~\ref{tab:bos-shared}).
Score masking redistributes attention to the remaining keys.
We test whether their increased contribution reconstructs the removed BOS contribution with the baseline
queries, keys, and values held fixed.

For action query \(i\) and head \(h\), let \(a_{ib}^{(h)}\)
denote the attention mass on BOS and \(v_b^{(h)}\) its value.
The renormalized combination of the non-BOS values is
\[
  \bar v_{i,-b}^{(h)}
  =
  \frac{\sum_{j\neq b}a_{ij}^{(h)}v_j^{(h)}}
       {1-a_{ib}^{(h)}}.
\]
Let \(W_O\) denote the linear output projection and \(g\)
the timestep-dependent gate.
Define the gated, projected BOS contribution as
\[
  B_i
  =
  g \odot W_O
  \left(
    \operatorname{concat}_{h}
    \big[a_{ib}^{(h)}v_b^{(h)}\big]
  \right),
\]
and the additional non-BOS contribution introduced by
renormalization as
\[
  C_i
  =
  g \odot W_O
  \left(
    \operatorname{concat}_{h}
    \big[a_{ib}^{(h)}\bar v_{i,-b}^{(h)}\big]
  \right).
\]
Value zeroing preserves the attention weights and removes
only the BOS contribution:
\begin{equation}
  \Delta o_{\mathrm{zero},i}=-B_i.
  \label{eq:vzero}
\end{equation}
Score masking also increases the contribution of the remaining
keys through renormalization:
\begin{equation}
  \Delta o_{\mathrm{mask},i}
  =
  C_i-B_i
  =
  \delta_i.
  \label{eq:mask-gap}
\end{equation}
Exact local reconstruction therefore requires \(C_i=B_i\)
after projection and gating; it does not require the underlying
value vectors to be identical.

We measure these local quantities in baseline forward passes.
To avoid numerically unstable renormalization, we exclude queries
with BOS mass above \(1-10^{-4}\) on any head.
This criterion retains \(97.2\%\) of queries overall, but only
approximately half at layer 1, where BOS attention is largest.
Averaged over eligible queries, layers, and denoising steps,
the mean \(\lVert\delta_i\rVert\) is \(5.67\), compared with
\(2.96\) for the mean \(\lVert B_i\rVert\).
The ratio of these means is \(1.91\).
Thus, on average over the retained queries, attention
renormalization does not nearly reconstruct the removed BOS
contribution: the local masking perturbation is larger,
not smaller, than the local value-zeroing perturbation.

These fixed-baseline measurements describe the immediate attention-sublayer perturbations, whereas \(D_{\mathrm{act}}\) measures the final action after subsequent layers and denoising steps.

\section{Closed-Loop BOS Interventions}
\label{app:bos-closed-loop}

\begin{figurehere}
  \centering
  \includegraphics[width=0.8\linewidth]{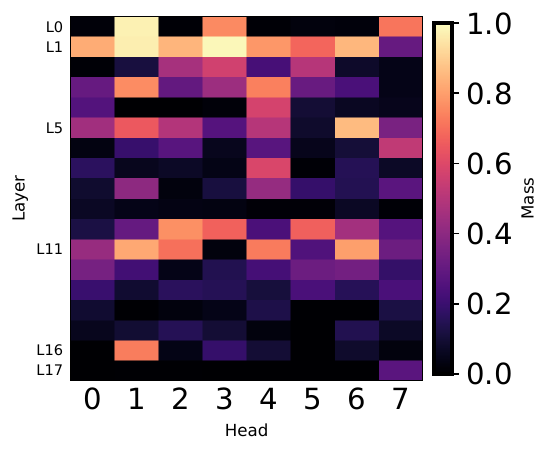}
  \caption{Per-head incoming mass on \texttt{<bos>}, by layer; Figure~\ref{fig:bos-layers} gives the head mean.}
  \label{fig:bos-heads}
\end{figurehere}

The BOS edits in Section~\ref{sec:bos-information} use the same execution protocol as Appendix~\ref{app:protocol}: 50 resets per task, seeds 0--49.
Score masking drops the \texttt{<bos>} key and renormalizes.
Value zero keeps the weights and sets the BOS value to zero.
Replacement keeps the weights and copies the entire BOS value from another task at seed 99.
Table~\ref{tab:bos-closed-loop} gives the success counts.
Table~\ref{tab:bos-closed-paired} gives the paired flips on the same 250 episodes.
The \(p\)-value is the exact two-sided McNemar test.
The mechanism of the value-zero increase is unresolved
(Section~\ref{sec:bos-information}).

\begin{tablehere}
\caption{Paired success changes for the BOS edits on the same 250 episodes.
\(s\!\to\!f\) is success to failure, and \(f\!\to\!s\) is failure to success.
The difference is in percentage points.
The \(p\)-value is an exact two-sided McNemar test.}
\label{tab:bos-closed-paired}
\centering
\footnotesize
\setlength{\tabcolsep}{4pt}
\begin{tabular}{@{}lrrrr@{}}
\toprule
Edit & \(s\!\to\!f\) & \(f\!\to\!s\) & Diff. & \(p\) \\
\midrule
Mask & 17 & 26 & \(+3.6\) & 0.22 \\
Value 0 & 17 & 50 & \(+13.2\) & \(6.7\times 10^{-5}\) \\
Replacement & 17 & 17 & 0 & 1 \\
\bottomrule
\end{tabular}
\end{tablehere}

On five reversed tasks, value zero and the unmodified BOS value were paired on the same seeds 0--49.
Value zero succeeded on 118 of 250 episodes and the unmodified condition on 99 of 250, a paired difference of \(+7.6\) points (6 successes became failures and 25 the reverse; \(p=8.8\times 10^{-4}\)).
The increase is not in CloseDrawer, where the unmodified run already succeeds on 46 of 50.
It comes from CloseCabinet (21 to 28), TurnOffMicrowave (1 to 8), and OpenFridge (11 to 16).
PickPlaceCabinetToCounter stays at 21 against 20.

\begin{tablehere}
\caption{Closed-loop success on the reversed tasks.
Value zero is paired with the unmodified BOS value on seeds 0--49.
Each cell is successes out of 50 episodes.}
\label{tab:bos-reversed}
\centering
\footnotesize
\setlength{\tabcolsep}{4pt}
\begin{tabular}{@{}lcc@{}}
\toprule
Task & Value 0 & Unmodified \\
\midrule
CloseDrawer & 45 & 46 \\
CloseCabinet & 28 & 21 \\
TurnOffMicrowave & 8 & 1 \\
OpenFridge & 16 & 11 \\
PickPlaceCabinetToCounter & 21 & 20 \\
\midrule
Total & 118 & 99 \\
\bottomrule
\end{tabular}
\end{tablehere}

A second run on the five tasks and seeds of Table~\ref{tab:bos-closed-loop} scales the BOS value, \(v'=\lambda v\), and leaves the attention weights and the key unchanged.
\(\lambda=1\) succeeds on 114 of 250 episodes, \(\lambda=0\) on 129 of 250 (\(+6.0\) points), and \(\lambda=0.5\) on 128 of 250 (\(+5.6\) points).
Of the 250 seeds, 94 succeed under both \(\lambda=0\) and \(\lambda=1\), 20 only at \(\lambda=1\), and 35 only at \(\lambda=0\); the exact two-sided McNemar \(p\) is \(0.058\), and the Wald 95\% confidence interval on the difference is \([+0.2,+11.8]\) points.
On seeds where both \(\lambda=0\) and \(\lambda=1\) succeed, value zero finishes in fewer environment steps: 393 against 443 over 94 pairs on these tasks (shorter in 71, longer in 23), and 280 against 334 over 93 pairs on the reversed tasks.
The gap is largest on CloseFridge, CloseCabinet, and OpenFridge.
The pick-and-place tasks take about the same number of steps.
The value-zero column of Table~\ref{tab:bos-reversed} is the same \(\lambda=0\) edit on the reversed tasks: 118 of 250 against 99 of 250.

\begin{tablehere}
\caption{Closed-loop success when the BOS value is scaled by \(\lambda\).
Each cell is successes out of 50 episodes.
Seeds 0--49.
\(\lambda=1\) is the unmodified value.}
\label{tab:bos-scale}
\centering
\footnotesize
\setlength{\tabcolsep}{4pt}
\begin{tabular}{@{}lccc@{}}
\toprule
Task & \(\lambda=0\) & \(\lambda=0.5\) & \(\lambda=1\) \\
\midrule
CloseFridge & 32 & 30 & 22 \\
OpenCabinet & 26 & 29 & 26 \\
OpenDrawer & 31 & 32 & 29 \\
PickPlace & 32 & 34 & 34 \\
TurnOnMicrowave & 8 & 3 & 3 \\
\midrule
Total & 129 & 128 & 114 \\
\bottomrule
\end{tabular}
\end{tablehere}

Table~\ref{tab:bos-closed-loop} is a separate execution of the unmodified policy, at 107 of 250; on a fixed observation, \(\lambda=1\) reproduces that pass with zero action difference.
Success labels differ on 27 of 250 seeds, 10 only in the Table~\ref{tab:bos-closed-loop} run and 17 only at \(\lambda=1\).
The seed fixes the environment generator and the diffusion noise but not the order of attention reductions on the GPU, and a small action difference can change the contact sequence.
Each gain is therefore the paired difference inside its own run: \(+13.2\) points from 107 for value zero in Table~\ref{tab:bos-closed-loop}, and \(+6.0\) points from 114 for \(\lambda=0\).
The unmodified totals across the three executions are 107, 113 (Table~\ref{tab:endpoint-success}), and 114.

\begin{tablehere}
\caption{Unmodified success in the two closed-loop runs, on the same seeds.
The first column is the baseline of Table~\ref{tab:bos-closed-loop}.
The second is \(\lambda=1\).
The last two columns count seeds whose success label disagrees.}
\label{tab:bos-rerun}
\centering
\footnotesize
\setlength{\tabcolsep}{3pt}
\begin{tabular}{@{}lcccc@{}}
\toprule
Task & Baseline & \(\lambda=1\) & Baseline only & \(\lambda=1\) only \\
\midrule
OpenDrawer & 28 & 29 & 1 & 2 \\
OpenCabinet & 23 & 26 & 3 & 6 \\
TurnOnMicrowave & 5 & 3 & 2 & 0 \\
CloseFridge & 19 & 22 & 2 & 5 \\
PickPlace & 32 & 34 & 2 & 4 \\
\midrule
Total & 107 & 114 & 10 & 17 \\
\bottomrule
\end{tabular}
\end{tablehere}

\section{Offset-Profile Routing Edits}
\label{app:offset-profile}

The layer-wise spread ratios, the uniform-routing depth breakdown, and the collapse figure below accompany Section~\ref{sec:mixing}.
The RMSE between two unmodified noises on one observation is \(0.333\) on the 24 observation--noise pairs of Table~\ref{tab:spread-ratio} and \(0.334\) in Table~\ref{tab:uniform-rmse}, a reference distance rather than \(D_{\mathrm{act}}\).
When the measured offset profile is the mean of the same unmodified action-key rows, the median cosine of the reconstructed rows is \(0.882\), and \(83\%\) of rows have cosine at least \(0.7\).
Agreement is weaker in the last layers: the mean cosine is \(0.625\) at layer 15 and \(0.619\) at layer 16.
In Tables~\ref{tab:uniform-spread} and~\ref{tab:uniform-rmse}, uniform routing of the unmodified attention gives final spread \(0.023\) and \(D_{\mathrm{act}}=0.452\).
The uniform row of Table~\ref{tab:offset-profile} is the leave-one-task-out result on the 24 observation--noise pairs of Table~\ref{tab:spread-ratio}: final spread \(0.024\) and \(D_{\mathrm{act}}=0.438\).

\begin{tablehere}
\caption{Spread ratio of each layer under unmodified attention.
The mean is over 24 observation--noise pairs and 10 denoising steps.
Layer 0 is the first layer, as in Table~\ref{tab:rope-pair}.}
\label{tab:spread-ratio}
\centering
\small
\begin{tabular}{lc@{\hspace{1.2em}}lc}
\toprule
Layer & Ratio & Layer & Ratio \\
\midrule
L0  & 1.014 & L9  & 1.001 \\
L1  & 1.000 & L10 & 1.006 \\
L2  & 1.000 & L11 & 1.015 \\
L3  & 0.999 & L12 & 1.012 \\
L4  & 0.998 & L13 & 1.012 \\
L5  & 0.999 & L14 & 1.025 \\
L6  & 1.001 & L15 & 1.006 \\
L7  & 0.994 & L16 & 1.011 \\
L8  & 1.007 & L17 & 1.040 \\
\bottomrule
\end{tabular}
\end{tablehere}

\begin{tablehere}
\caption{Final action-token spread after uniform action--action routing, divided by the paired baseline.
Layer numbers are 0-based, as in Table~\ref{tab:spread-ratio}.}
\label{tab:uniform-spread}
\centering
\small
\begin{tabular}{lc}
\toprule
Condition & Final spread \\
\midrule
baseline & 1.000 \\
layer 11 & 0.768 \\
layer 17, lowest & 0.592 \\
layer 7, highest & 1.007 \\
\midrule
layers 0--5 & 0.790 \\
layers 6--11 & 0.170 \\
layers 12--17 & 0.479 \\
all 18 layers & \textbf{0.023} \\
\bottomrule
\end{tabular}
\end{tablehere}

\begin{tablehere}
\caption{\(D_{\mathrm{act}}\) after the same uniform action--action routing.
The RMSE between two unmodified noises on one observation is \(0.334\), a reference distance rather than \(D_{\mathrm{act}}\).}
\label{tab:uniform-rmse}
\centering
\small
\begin{tabular}{lc}
\toprule
Condition & \(D_{\mathrm{act}}\) \\
\midrule
layer 0, lowest & 0.001 \\
layer 11 & 0.162 \\
layer 17, highest & 0.201 \\
\midrule
layers 0--5 & 0.119 \\
layers 6--11 & \textbf{0.388} \\
layers 12--17 & 0.295 \\
all 18 layers & \textbf{0.452} \\
\bottomrule
\end{tabular}
\end{tablehere}

\begin{figurehere}
  \centering
  \includegraphics[width=\linewidth]{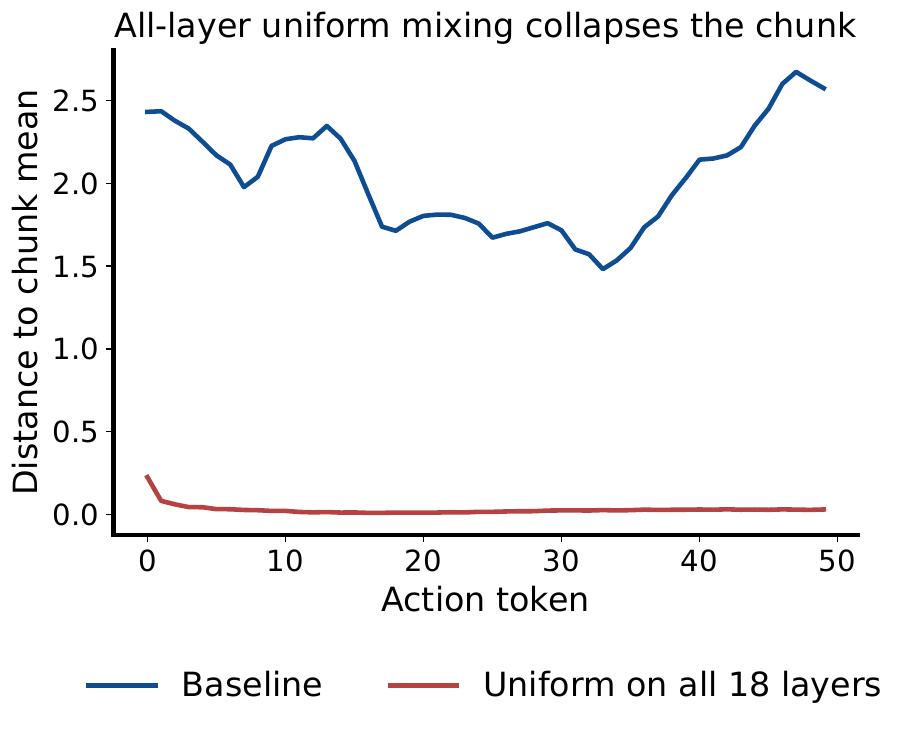}
  \caption{Absolute distance of each action token to the chunk mean.
  This is not the final-spread ratio in Table~\ref{tab:uniform-spread}.
  A large distance means the token still differs from the rest of the chunk.
  A distance near zero means the tokens have been pulled together.
  Blue is unmodified attention.
  Red is all-layer uniform action--action routing.
  Both curves average the 24 observation--noise pairs of Table~\ref{tab:spread-ratio}.
  Single-layer and block results are in Tables~\ref{tab:uniform-rmse} and~\ref{tab:uniform-spread}.}
  \label{fig:uniform-collapse}
\end{figurehere}

Section~\ref{sec:structured-mixing} compares the four edits of Section~\ref{sec:mixing-intervention} on the 24 observation--noise pairs of Table~\ref{tab:spread-ratio}: OpenDrawer, TurnOnMicrowave, and CloseFridge, environment seeds 0 and 1, states 0 and 25, and noise seeds 0 and 1.
For each held-out task, \(T_{\ell,h}\) is estimated on the other two tasks, separately for each layer and each of the 8 heads, and applied on that held-out task.
The profile and the edited weights are Equations~\eqref{eq:offset-profile} and~\eqref{eq:offset-routing}.
Rows with action-key mass below \(10^{-8}\) are left out of the estimate of \(T_{\ell,h}\).
The edit then renormalizes each query over the offsets it can reach, and offset \(+50\) is never read.
Uniform routing on all 18 layers in Table~\ref{tab:offset-profile} gives final spread \(0.024\) and \(D_{\mathrm{act}}=0.438\).
In Tables~\ref{tab:uniform-spread} and~\ref{tab:uniform-rmse}, the same uniform routing on these pairs gives final spread \(0.023\) and \(D_{\mathrm{act}}=0.452\).

The cosine compares \(p_i\) with the row reconstructed from \(T_{\ell,h}(j-i)\), where \(T_{\ell,h}\) is the mean of the same unmodified action-key rows.
Of \(1{,}727{,}975\) retained rows, the mean cosine is \(0.831\), the 10th, 50th, and 90th percentiles are \(0.610\), \(0.882\), and \(0.971\), and the fractions at or above \(0.5\) and \(0.7\) are \(0.950\) and \(0.832\).
Mean total variation is \(0.284\).
Table~\ref{tab:offset-cosine} gives the mean cosine by layer.

\begin{tablehere}
\caption{Mean cosine between each unmodified action-key row and the row read from \(T_{\ell,h}(j-i)\).
Layer 0 is the first layer.}
\label{tab:offset-cosine}
\centering
\small
\begin{tabular}{lc@{\hspace{1.2em}}lc}
\toprule
Layer & Cosine & Layer & Cosine \\
\midrule
L0  & 0.928 & L9  & 0.829 \\
L1  & 0.900 & L10 & 0.875 \\
L2  & 0.950 & L11 & 0.902 \\
L3  & 0.856 & L12 & 0.734 \\
L4  & 0.897 & L13 & 0.775 \\
L5  & 0.939 & L14 & 0.777 \\
L6  & 0.866 & L15 & 0.625 \\
L7  & 0.824 & L16 & 0.619 \\
L8  & 0.893 & L17 & 0.768 \\
\bottomrule
\end{tabular}
\end{tablehere}

Offset shuffle uses seeds \(0,\ldots,4\) and key permutation uses seeds \(1000,\ldots,1004\), five permutations each, as defined in Section~\ref{sec:mixing-intervention}.
The mean entropy of the shuffled 50-key rows is \(3.393\)--\(3.414\) nats, against \(3.512\) nats for shared offset routing.
Key permutation leaves that row entropy at \(3.512\) nats.
Table~\ref{tab:offset-seeds} lists every permutation on all 18 layers.

\begin{tablehere}
\caption{Final spread and \(D_{\mathrm{act}}\) for each permutation.
Spread is divided by the paired baseline.
The block edits all 18 layers.}
\label{tab:offset-seeds}
\centering
\small
\begin{tabular}{lcccc}
\toprule
& \multicolumn{2}{c}{Offset shuffle} & \multicolumn{2}{c}{Key permutation} \\
\cmidrule(lr){2-3}\cmidrule(lr){4-5}
Seed & Spread & \(D_{\mathrm{act}}\) & Spread & \(D_{\mathrm{act}}\) \\
\midrule
\multicolumn{5}{l}{All 18 layers} \\
0 & 0.306 & 0.420 & 0.067 & 0.425 \\
1 & 0.172 & 0.424 & 0.125 & 0.431 \\
2 & 0.194 & 0.420 & 0.130 & 0.421 \\
3 & 0.143 & 0.440 & 0.097 & 0.432 \\
4 & 0.154 & 0.432 & 0.058 & 0.440 \\
\bottomrule
\end{tabular}
\end{tablehere}

\paragraph{BOS edits and spread.}
Table~\ref{tab:bos-spread} gives the BOS edits of Section~\ref{sec:mixing} on the 24 observation--noise pairs.
Under BOS value zeroing the 24 paired spread ratios lie between \(1.219\) and \(1.651\).
What changes is the final denoised chunk, not the product of the per-layer residual spread ratios, which moves only from \(1.147\) to \(1.160\); the unmodified per-layer ratios match the earlier mixing run to within \(8.5\times10^{-5}\).
The score-mask row of Table~\ref{tab:bos-shared} is a different sample of 15 states: median \(D_{\mathrm{act}}=0.036\), mean \(0.051\).

\begin{tablehere}
\caption{
Mean final-chunk spread ratio and mean \(D_{\mathrm{act}}\)
after edits to the prefix \texttt{<bos>} token.
The 24 observation--noise pairs are OpenDrawer,
TurnOnMicrowave, and CloseFridge, environment seeds 0 and 1,
states 0 and 25, and two noises.
\(D_{\mathrm{act}}\) is Equation~\eqref{eq:d-act}.
Table~\ref{tab:bos-shared} is a different sample of 15 states.
}
\label{tab:bos-spread}
\centering
\small
\begin{tabular}{lcc}
\toprule
Condition & $r_{\mathrm{spread}}$ & $D_{\mathrm{act}}$ \\
\midrule
baseline & 1.000 & 0.000 \\
BOS score mask & 0.987 & 0.039 \\
BOS value zero & 1.381 & 0.235 \\
\bottomrule
\end{tabular}
\end{tablehere}

\paragraph{Closed-loop routing.}
Table~\ref{tab:route-closed-loop} uses the same edits on OpenDrawer, TurnOnMicrowave, and CloseFridge, 50 episodes each, with each task's profile estimated on the other two (offset shuffle seed 0, key permutation seed 1000).
Per task, shared offset routing changes success by \(+8\), \(+2\), and \(+2\) points (\(p\geq0.39\)).
The three other edits lower OpenDrawer by 58 points (\(p=3.7\times10^{-9}\)) and CloseFridge by 38 to 40 points (\(p\leq6.6\times10^{-5}\)); TurnOnMicrowave falls from 4 to 0.
Failed episodes run to the time limit rather than ending early.
A profile pooled over these three tasks and applied to OpenCabinet and PickPlaceCounterToCabinet lowers shared offset routing from 23 to 14 and from 34 to 21 successes, so the profile is estimated per task family; Table~\ref{tab:route-closed-loop} reports the three leave-one-task-out tasks.

\section{A SmolVLA Probe}
\label{app:smolvla}

We tested whether the three signatures identified in the \(\pi_{0.5}\) action expert transfer to SmolVLA~\cite{shukor2025smolvla}.
The probe used four synthetic observations, two noise samples per observation, ten denoising steps, and an additional all-zero-image control.
SmolVLA contains 16 expert layers.
Action--action attention in the tested layers is causally masked, while the remaining layers condition action queries on prefix representations.
Such causal action mixing is a defining architectural feature of SmolVLA.
All comparisons used paired observations and noises, and no closed-loop evaluation was performed.

\subsection{RoPE and Prefix Tokens}

We found no evidence for the radial RoPE mechanism observed in \(\pi_{0.5}\).
Under causal attention, the two action endpoints are not visibility-symmetric: the final action key is unavailable to earlier queries, and the causal-uniform edge-mass baseline is \(0.0625\).
Across the tested layers and denoising states, normal RoPE placed edge mass at or below this baseline.
Replacing RoPE with identity increased rather than removed the edge mass, and rolling positions did not relocate attention to the rolled geometric columns.
Disabling RoPE in layers 0--11 likewise did not eliminate the L12 deviation from identity.
SmolVLA also introduces no BOS token.
Its most-attended prefix position was instead the projected state token.
Score masking and value removal strongly changed the action output, while cross-example value replacement produced an action deviation comparable to changing the sampling noise.
We therefore interpret it as an observation-dependent proprioceptive channel rather than a token-matched analogue of the shared \texttt{<bos>} bias.

\subsection{Structured Action Routing}

We repeated the routing intervention while preserving SmolVLA's original causal support.
For action query \(i\), its total action-attention mass \(m_i^{\mathrm{act}}\) was redistributed uniformly only over keys \(j\le i\).
Future-action mass remained zero, and all prefix columns were unchanged:
\begin{equation}
  A'_{ij}=
  \begin{cases}
    m_i^{\mathrm{act}}/(i+1), & j\le i,\\
    0, & j>i.
  \end{cases}
  \label{eq:smolvla-causal-uniform}
\end{equation}

Under this causal-support-preserving edit, uniformizing every action-mixing layer left action-token spread at 45.4\% of baseline, and individual or partial-layer edits retained 78.8--88.4\% (Table~\ref{tab:smolvla-routing}).
The resulting \(D_{\mathrm{act}}\) did not reproduce the stronger drop found in \(\pi_{0.5}\).

\begin{tablehere}
\caption{Causal-support routing intervention on SmolVLA.
\(D_{\mathrm{act}}\) is the final chunk against the paired baseline (Equation~\eqref{eq:d-act}).
Relative final spread is the final action-token spread divided by that baseline.
Middle even layers are 4, 6, 8, and 10.
All even layers are \(0,2,\ldots,14\).
The RMSE between two unmodified noises on one observation is \(0.478\), a reference distance rather than \(D_{\mathrm{act}}\).}
\label{tab:smolvla-routing}
\centering
\small
\begin{tabular}{lcc}
\toprule
Intervened layers & \(D_{\mathrm{act}}\) & Relative final spread \\
\midrule
L0 only & 0.109 & 0.788 \\
L12 only & 0.083 & 0.884 \\
Middle even layers & 0.116 & 0.797 \\
All even layers & 0.307 & 0.454 \\
\bottomrule
\end{tabular}
\end{tablehere}

No tested \(D_{\mathrm{act}}\) exceeded \(0.478\), the RMSE between two unmodified noises on one observation, and single-layer edits changed the action by less than the all-layer edit.
The probe uses eight synthetic observation--noise pairs.

\end{document}